\documentclass[sigconf]{acmart}

\usepackage{subfigure}
\usepackage{lipsum}

\usepackage{tabularx}
\usepackage{wrapfig}
\usepackage{graphicx}
\usepackage{booktabs}
\usepackage{multirow}
\usepackage{array}
\usepackage{pifont}

\usepackage{balance}

\def\DatasetName{\textsc{UserPrefSum}}

\newcolumntype{P}[1]{>{\centering\arraybackslash}p{#1}}

\copyrightyear{2024}
\acmYear{2024}
\setcopyright{rightsretained}
\acmConference[CIKM '24]{Proceedings of the 33rd ACM International Conference on Information and Knowledge Management}{October 21--25, 2024}{Boise, ID, USA}
\acmBooktitle{Proceedings of the 33rd ACM International Conference on Information and Knowledge Management (CIKM '24), October 21--25, 2024, Boise, ID, USA}
\acmDOI{10.1145/3627673.3680011}
\acmISBN{979-8-4007-0436-9/24/10}

\makeatletter
\gdef\@copyrightpermission{
  \begin{minipage}{0.3\columnwidth}
{\includegraphics[width=0.90\textwidth]{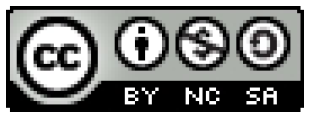}}
  \end{minipage}\hfill
  \begin{minipage}{0.7\columnwidth}
   \href{https://creativecommons.org/licenses/by-nc-sa/4.0/}{This work is licensed under a Creative Commons Attribution-NonCommercial-ShareAlike International 4.0 License.}
  \end{minipage}
  \vspace{5pt}
}
\makeatother
\begin{document}

\title{Personalized Video Summarization by Multimodal Video Understanding}

\author{Brian Chen}
\orcid{0000-0001-9025-0107}
\affiliation{%
  \institution{Samsung Research America}
  \department{VDIL}
  \city{Irvine}
  \state{California}
  \country{USA}
}
\email{b.chen1@samsung.com}

\author{Xiangyuan Zhao}
\orcid{0009-0009-8220-6347}
\affiliation{%
  \institution{Samsung Research America}
  \department{VDIL}
  \city{Irvine}
  \state{California}
  \country{USA}
}
\email{xiangyuan.z@samsung.com}

\author{Yingnan Zhu}
\orcid{0009-0000-5709-1335}
\affiliation{%
  \institution{Samsung Research America}
  \department{VDIL}
  \city{Irvine}
  \state{California}
  \country{USA}
}
\email{yingnan.z@samsung.com}

\renewcommand{\shortauthors}{Brian Chen, Xiangyuan Zhao, \& Yingnan Zhu}

\newcolumntype{P}[1]{>{\centering\arraybackslash}p{#1}}

\begin{abstract}
Video summarization techniques have been proven to improve the overall user experience when it comes to accessing and comprehending video content. If the user's preference is known, video summarization can identify significant information or relevant content from an input video, aiding them in obtaining the necessary information or determining their interest in watching the original video. Adapting video summarization to various types of video and user preferences requires significant training data and expensive human labeling.
To facilitate such research, we proposed a new benchmark for video summarization that captures various user preferences. Also, we present a pipeline called Video Summarization with Language (VSL) for user-preferred video summarization that is based on pre-trained visual language models (VLMs) to avoid the need to train a video summarization system on a large training dataset. The pipeline takes both video and closed captioning as input and performs semantic analysis at the scene level by converting video frames into text. Subsequently, the user's genre preference was used as the basis for selecting the pertinent textual scenes. The experimental results demonstrate that our proposed pipeline outperforms current state-of-the-art unsupervised video summarization models. We show that our method is more adaptable across different datasets compared to supervised query-based video summarization models. In the end, the runtime analysis demonstrates that our pipeline is more suitable for practical use when scaling up the number of user preferences and videos.
\end{abstract}




\maketitle

\section{Introduction}
\label{sec:intro}


The availability of online content, such as news articles, live broadcasts, and video blogs, has led to a demand for video summarization in different practical scenarios. The growing interest in multimodal learning has focused on the development of personalized video summaries using natural language queries \cite{momentdetr}. Unlike conventional video summarization methods \cite{CASum22} that solely rely on video content to capture repetitive scenes as highlights, query-guided video summarization \cite{lin2023univtg} incorporates information from natural language queries to produce concise video summaries. This approach provides users with condensed information, which is particularly useful for lengthy videos such as live streams and product reviews, where redundant content is often present.

\begin{figure}
    \centering
    \includegraphics[width=0.48\textwidth]{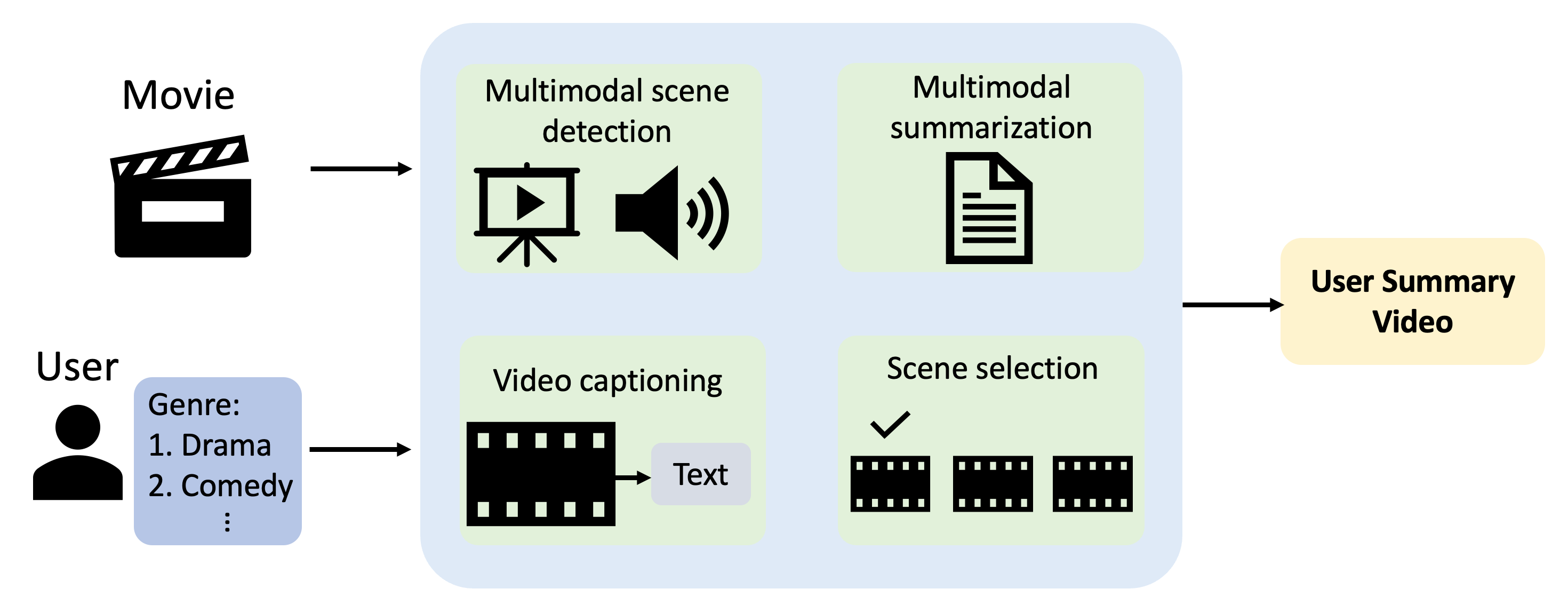}
    \caption{The overall structure of VSL. 
    VSL utilizes a captioning model to transform long videos into text and summarize the text to represent the video.}
    \label{fig:overall_VSL}
    \vspace{-0.3cm}
\end{figure}

Existing methods for query-guided video summarization \cite{momentdetr} have utilized text information but can only generate summaries based on descriptive sentences, such as "A shark is swimming under water," rather than more abstract concepts like genres, such as "Sci-Fi," "Romance," or "Comedy." However, genre tags can be easily obtained from recommendation systems and can represent user preferences when recommending videos. Unfortunately, current query-guided models struggle to understand genre-specific tags due to limitations in the training data, which was only trained with paired natural language sentences and lacks a deeper understanding of videos. Recently, there have been studies exploring user-preferred video summarization \cite{momentdetr,qddetr,lin2023univtg}, which aim to generate summaries based on specific user applications. However, genre-specific user preferences have not been thoroughly explored. In this work, our goal is to generate video summaries that are conditioned on user preferences, resulting in user-centric video summarization.

In order to facilitate research on long-video summarization with user-preferred information, we have collected a large-scale video dataset called \DatasetName. This dataset consists of over 1K movie videos from Condensed Movies \cite{condensed_movie}, covering a diverse range of genres (21 classes in total). For genre-based video summarization, users can provide one or multiple preferred movie genres as a query. We start by detecting scenes in the videos and then use the zero-shot ability of CLIP\cite{CLIP} to automatically label the genre of each scene. The final video summarization contains multiple scenes that are related to the given genre. This approach allows for a more realistic and user-centered evaluation of video summarization methods. 
This is motivated by the fact that video summarization can be quite subjective. Instead, we are offering a more unbiased approach to assess such situations by representing user preference by genres. This is based on our hypothesis on the Content-Based Recommendation System \cite{pazzani2007content}, the underlying principle of this type of recommendation system being that if a user enjoyed a certain movie or show, they may also enjoy something similar. 


Motivated by the observations mentioned above, we propose a Video Summarization with Language (VSL) approach, as depicted in Figure \ref{fig:overall_VSL}. VSL is based on the Socratic model \cite{zeng2022socratic} and consists of four components: multimodal scene detection, video captioning, multimodal summarization, and video selection. The multimodal scene detection component utilizes both video and closed captioning as inputs to identify clear scene cuts that preserve the integrity of both the video and dialogue in the input movie. Video captioning converts the input movie from the video domains to a text-based captioning domain for semantic analysis. The multimodal summarization component generates summaries for both video captioning and closed captioning, capturing the most important information at the scene level. The video selection component analyzes the summarization results from the multimodal summarization component, taking into account the input genre, in order to select appropriate scenes and create the final summary video that aligns with the specific user preferences. Experimental results demonstrate that VSL outperforms current state-of-the-art methods in both general video summarization (TVSum\cite{song2015tvsum}) and user-specific video summarization (\DatasetName). We also test summarization in user-generated videos SumMe\cite{summe}. Additionally, we conduct runtime analysis on varying numbers of videos and user preferences to showcase the practical applicability of the VSL model.


To summarize, our contributions include:
\begin{itemize}
\vspace{-0.3cm}
\item A new benchmark is proposed to facilitate the research of user-preferred video summarization.
    \item A language-based architecture VSL makes use of pre-trained large language models for video captioning and multimodal summarization. This allows the architecture to analyze natural language at a semantic level, which takes into account multimodal input and genre recommendation, an aspect overlooked by previous studies.
\item A runtime analysis has been conducted to demonstrate the scalability of our model. Unlike the previous state-of-the-art method that requires feeding user preference queries to each video individually, our VSL can generate summarization in parallel. 
This real-time capability is crucial for practical applications.

\end{itemize}

\section{Related work}
\label{sec:related_work}


\subsection{General video summarization}
\noindent \textbf{Datasets.} General video summarization, also known as Query-agnostic summarization, involves generating a concise version of a given video without any user query. Datasets such as those mentioned in \cite{summe, song2015tvsum, mahasseni2017unsupervised, jiang2022joint} are based solely on visual cues.

\noindent \textbf{Method.} The advances in unsupervised learning have demonstrated promising potential in general video summarization. A notable example is the CA-SUM\cite{CASum22} approach, which incorporates length regularization loss and concentrated attention to achieve high F-scores on both the TVSum\cite{song2015tvsum} and SumMe\cite{summe} datasets, outperforming GAN-based models\cite{ac_sum}. Unlike supervised query-guided video summarization, CA-SUM is an unsupervised model that also offers the possibility of zero-shot capability. However, a limitation of CA-SUM is its reliance on a pre-defined attention mechanism, which restricts its ability to perform semantic analysis on complex video content, particularly in the case of movies.

\subsection{User specific video summarization.}
\noindent \textbf{Datasets.}
The primary objective of query-guided video summarization is to allow users to customize the summary by specifying specific text keywords (e.g., "A man riding a horse"). This practical approach is the main focus of our research. Another relevant task, called Moment retrieval, involves locating specific moments in a video using a natural language query. Several datasets \cite{sharghi2017query, nalla2020watch, wu2022intentvizor} have been proposed or modified for this purpose.
While some of these datasets only provide one moment for each query-video pair, others offer multiple moments. Unlike previous datasets that primarily focus on retrieving relative moments using natural language queries, our focus is on retrieving moments/scenes based on user preferences. To construct such a user preference for evaluation, we utilize genre tags to represent the user's preference and retrieve multiple moments for our final video summarization.

\noindent \textbf{Method.}
Several approaches \cite{xiao2020convolutional,huang2020query,xiao2020query,narasimhan2021clip,vasudevan2017query} have been developed to tackle video summarization/moment retrieval tasks. Some of these approaches focus on scoring generated moment proposals, predicting moment start-end indices, or regressing moment coordinates. However, these approaches require manual preprocessing steps (e.g., proposal generation) or postprocessing steps (e.g., non-maximum suppression) that are not trainable end-to-end. Another line of work, represented by DETR-based methods \cite{momentdetr,qddetr,lin2023univtg}, treats video summarization/moment retrieval as a direct set prediction problem. These methods take video and user query representations as inputs and directly output moment coordinates and saliency scores.
However, existing approaches often generate summaries using natural language sentences, which can be costly in terms of describing user preferences. 
Furthermore, as the number of users increases, the user preference information can become extensive. To address these challenges, we propose a method called Video Summarization with Language (VSL), which leverages a video-to-text model to process all information in the text space. Our method demonstrates the ability to handle diverse user preferences without increasing the computation time. Also, due to the design nature of our model, it can easily incorporate other user-preference information besides genre.


\begin{figure*}
    \centering
    \includegraphics[width=\textwidth]{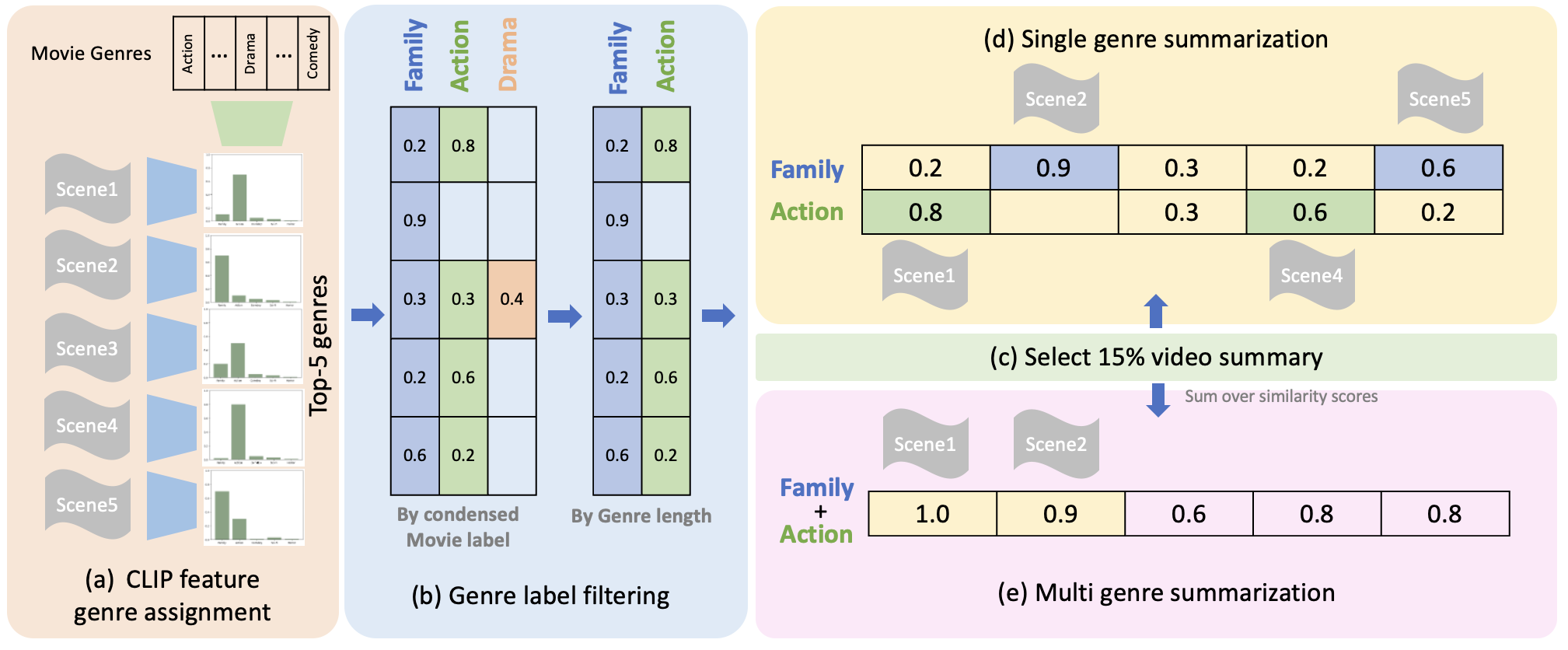}
    \caption{\textbf{Dataset creation pipeline.} (a) We employ the CLIP features to compute the similarity between the genre to each frame in zero-shot manner and then aggregate the results to obtain the distribution of genre labels at the scene level.
(b) We filter the genre labels based on the movie genre annotations provided by GT and a threshold for the length of the genre-specific summarization.
(c) We selected the top 15\% of scenes with the highest confidence score for a particular genre to include in the video summarization.
(d) For each genre query, we created a video summarization that focuses on scenes belonging to that genre.
(e) In the multi-genre summarization, we combine the confidence scores across different genres to determine the overall confidence score.}
    \label{fig:dataset}
\end{figure*}

\section{Dataset}
\label{sec:dataset}

To address the task of user-preferred video summarization, we have developed a labeled dataset called \DatasetName, which is designed specifically for movie summarization. In the following sections, we will describe the procedure for automatic labeling using CLIP \cite{CLIP} and the approach for filtering the labels. The process of creating the data is illustrated in Figure \ref{fig:dataset}. Our dataset is on top of the public dataset Condensed Movies \cite{condensed_movie}. We will release the code and dataset annotation for reproducibility.

\subsection{Scene segmentation for summarization}
The process of video summarization involves dividing the video into multiple scenes, each of which has a similar semantic meaning. In this context, we assume that scenes sharing the same semantic similarity belong to the same movie genre. To achieve this, we follow the approach proposed by Bose et al. \cite{bose2023movieclip} and utilize PySceneDetect\footnote{https://pyscenedetect.readthedocs.io/en/latest/} to segment the movie clips in the Condensed Movies \cite{condensed_movie} into multiple scenes.

\subsection{CLIP-based movie genre labeling}
In this section, we describe how CLIP \cite{CLIP} is used to associate genre labels with specific segments of movie scenes from the previous section, as illustrated in Figure \ref{fig:dataset}(a). To alleviate the burden of manually annotating long videos in \DatasetName, we leverage the zero-shot capabilities of CLIP, following the approach proposed in \cite{bose2023movieclip}, to tag the scenes. A notable difference is that we assign movie genres as labels to the scenes, rather than background scene labels. The genre labels are obtained from the Moviescope dataset \cite{2019Moviescope}, which is a subset of the Condensed Movies \cite{condensed_movie} genre. This dataset comprises 21 movie genre types, listed in Table \ref{tab:sota}. By employing the CLIP prompt engineering technique, we use a specific prompt for each movie genre, such as "A photo of a {label}, a type of movie genre." This prompt allows us to assign labels from our movie genre list to individual frames or scenes in video clips. Given a movie clip, we generate multiple scene segments, denoted as $s_{i}$, following the approach described in the previous section, where $(i=1,...,N)$. Let's assume that scene $s_i$ contains $T$ frames. We utilize CLIP's visual encoder to extract frame-wise visual embeddings $v_{t}$, where $(t=1,...,T)$. For each movie genre, we employ CLIP's text encoder to extract embeddings $g_{l}$, where $(l=1,2,...,L)$, corresponding to the genre-specific prompts. To compute the similarity score matrix $M$, whose entries $M_{lt}$ represent the similarity between the genre-specific prompt $g_{l}$ and the frame-wise visual embedding $v_{t}$, we use the following equation: \begin{equation}\label{simmatrixscore} M_{lt}=\frac{g_{l}^{T}v_{t}}{\lVert g_{l} \rVert_{2} \lVert v_{t} \rVert_{2}} \end{equation} 
To obtain an aggregate score for each scene label, we perform a temporal average pooling over the similarity matrix $M_{lt}$ following \cite{luo2022clip4clip} to capture temporal information. This provides us with the genre score $G_{l}$ for a given scene, as the visual content within a scene tends to remain relatively consistent. Here, $G_{l}$ represents the distribution of confidence scores for each genre assigned to a scene. Finally, we select the top 5 genre labels, as shown in Figure \ref{fig:dataset}(a).

\begin{figure*}
    \centering
    \includegraphics[width=\textwidth]{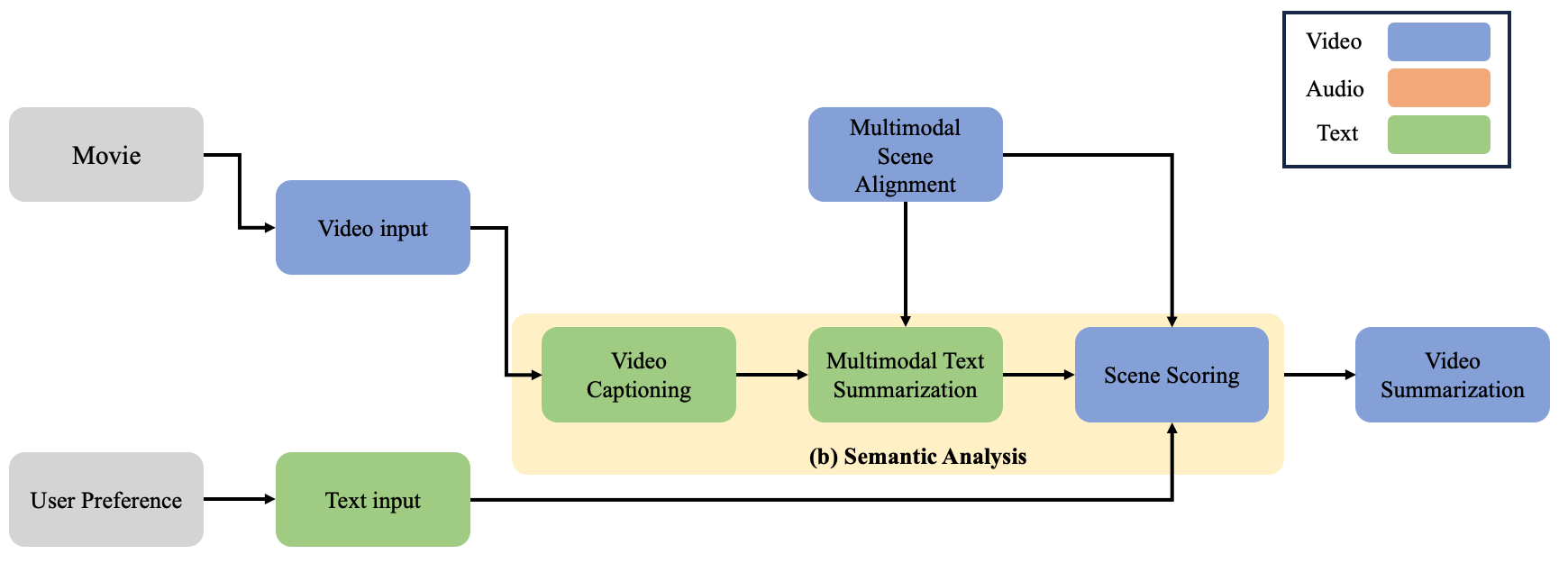}
    \caption{The detailed structure for VSL consists of three main steps. (a) In the first step, the input movie's audio and video backbones undergo processing through multimodal scene detection. (b) In the second step, semantic analysis is performed, which generates a score for each scene based on the results of multimodal scene detection. (c) The final step involves video summarization, where the scenes with the highest scores are selected to generate the summary video.}
    \label{fig:pipeline}
\end{figure*}

Furthermore, we make use of the genre labels assigned to each scene at the movie level in Condensed Movies \cite{condensed_movie} to exclude genre labels that are not annotated for a particular movie, as illustrated in Figure\ref{fig:dataset}(b). As a result, each scene $s$ in the movie is associated with a maximum of 3 genres based on the ground truth annotation. In the given example, only 3 genres ('Family,' 'Action,' and 'Drama') were retained since these were the genres annotated in the movie.



\begin{figure}
    \centering
    \includegraphics[width=0.48\textwidth]{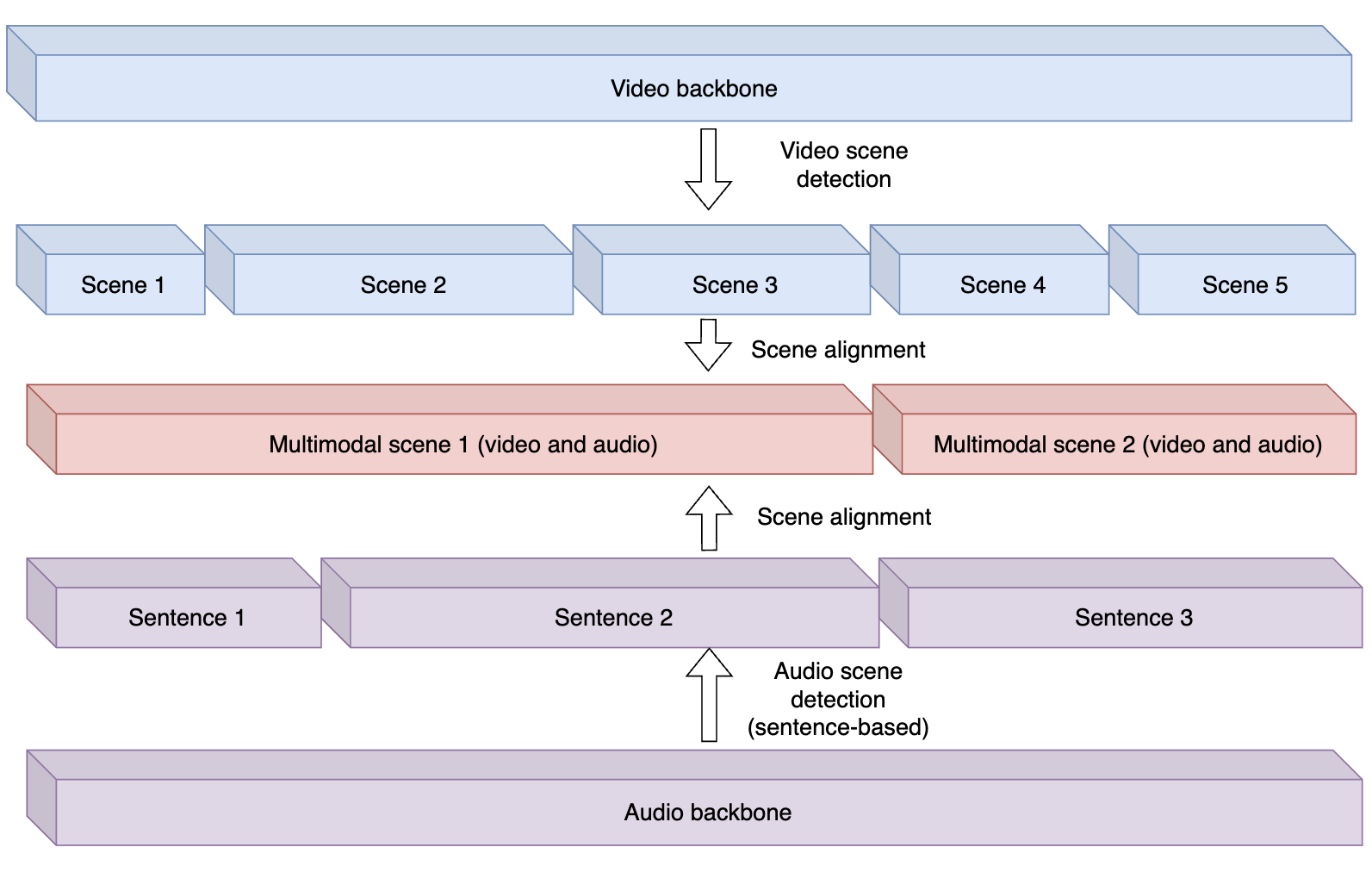}
    \caption{The process of multimodal scene detection involves the use of separate algorithms for video and audio backbones. These backbones analyze the input data independently to detect scenes. The resulting scene detection outcomes from each backbone are then aligned and combined at the frame level to obtain the final multimodal scene detection results.}
    \label{fig:multimodal_scene_detection}
\end{figure}


\subsection{Multi-genre label creation}
In order to accommodate users with multiple genre preferences, we aggregate the $G_l$ values for different genres and apply the same filtering strategy described in the previous sections. This approach will result in a video summarization that is 15\% of the original length (following \cite{song2015tvsum}) when the user provides 2 or 3 genre queries, as depicted in Figure \ref{fig:dataset}(e). The selected scenes will be the ground truth label of the video summarization. The idea behind this is to create a final video summary that includes a combination of different genres. We evaluate the performance of this approach in the ablation study described in Section \ref{sec:more_ex_multi}.

\subsection{Dataset analysis}

We gathered a total of 1800 queries related to 9000 moments across 1008 videos. Unlike other moment retrieval datasets, \DatasetName~can have multiple separate moments paired with a single query. On average, there are 9 moments per query in a video. This is in contrast to previous datasets where there are fewer than two moments per query. 
One key difference is that our queries represent a single-word high-level semantic genre, while previous studies focused on using natural language queries to retrieve more specific video segments.
To evaluate the quality of the automatic labeling process, a human evaluation was conducted by randomly sampling 10\% of the annotations. The Inter-Annotator Agreement (IAA) \cite{artstein2017inter} was 74.3\%, approaching the accuracy of human annotation.

\section{Method}
\label{sec:method}

In this section, we present the VSL architecture as in Figure~\ref{fig:pipeline}. The goal of VSL is to identify the most significant scenes in a given movie based on the input genres. VSL offers two modes for video summarization: general video summarization and user-specific video summarization. In general mode, the input genres are extracted from the movie metadata, representing its inherent genres. In the user-specific mode, the input genres are obtained from an upstream recommendation system, reflecting the user's preferences.
To execute the VSL architecture, the first step involves dividing the input movie into separate video and audio backbones. Next, a scene detection process is performed to convert the input movie into a sequential scene set $\{ s_{1}, ..., s_{i}\}$, where $i$ ranges from 1 to $N$. Subsequently, semantic analysis is applied to all scenes according to input genres, resulting in a list of saliency scores $\{M_{1}, ..., M_{i}\}$. Finally, the process of generating a summary video involves selecting scenes with the highest saliency scores. Compared to previous approaches, VSL improves the scene detection algorithm and offers capabilities in semantic analysis. Further details will be explained in the following sections.

\subsection{Multimodal scene detection}

The objective of scene detection is to identify frames in a video where significant semantic changes occur. Previous studies, such as MovieCLIPs \cite{bose2023movieclip}, have typically employed video-based algorithms for scene detection. However, this approach often results in cutting the video in the middle of a dialogue, which can create a confusing summary video for users. To address this problem, we propose a multimodal scene detection approach in our architecture. The multimodal scene detection method incorporates both video and audio signals as inputs, ensuring that the generated scenes maintain the integrity of both the video and the dialogue.

Figure \ref{fig:multimodal_scene_detection} illustrates the process of our scene detection, which involves two different detections. The first, known as video-based scene detection, is responsible for identifying shot changes in the input video. On the other hand, the audio-based scene detection generates closed captioning and subsequently aggregates and segments them at the sentence level. To align the results of both video and audio scene detections on a frame level, a multimodal scene alignment module is employed by combining the boundaries of both scenes and audio to establish a new boundary. Implementation is done by tracking start and end times across both modalities and then merging the boundaries based on the union of the two. 
This alignment allows multimodal scene detection to accurately determine the changing points in the input movie, allowing the movie to be split into scenes while preserving the integrity of both the video and dialogue.

\begin{table*}[h]
\centering
\resizebox{\textwidth}{!}{
\begin{tabular}{ccccccccccccccc}
\toprule
\textbf{Model} & \textbf{Overall} & \textbf{Ac} & \textbf{Ani} & \textbf{Bio} & \textbf{Com} & \textbf{Cri} & \textbf{Drm} & \textbf{Fmy} & \textbf{Fntsy} & \textbf{Hrrr} & \textbf{Myst} & \textbf{Rom} & \textbf{ScF} & \textbf{Thrl} \\ \midrule
MomentDETR \cite{momentdetr}           & 22.5             & 20.5        & 22.0         & 18.1           & 21.7         & 20.7           & 22.1           & 21.9         & 22.3           & 24.0          & 23.8          & 24.5         & 26.1         & 22.7          \\ 
QD-DETR  \cite{qddetr}       &    23.1        &   \textbf{21.4}     & 24.1         & 13.1          & 22.4         & 20.7         & 21.2         & 24.3         & 28.4           & 23.1          & 25.1          & 25.4         & 20.3         & 21.5          \\ 
UniVTG \cite{lin2023univtg}      & 23.6             & 20.8        & \textbf{26.4}         & 25.5         & 23.7         & 21.8         & 24.1           & 23.1           & \textbf{31.9}           & 18.3          & 27.1          & 24.5         & 24.8         & 23.7            \\ \midrule
\textbf{VSL}  & \textbf{26.8}             & 21.1        & 24.5         & \textbf{34.2}         & \textbf{24.9}         & \textbf{26.3}         & \textbf{27.3}         & \textbf{35.5}         & 31.8           & \textbf{30.5}          & \textbf{27.4}          & \textbf{37.7}         & \textbf{26.7}           & \textbf{27.3}          \\ \bottomrule
\end{tabular}
}
\vspace{2mm}
\caption{F1 score for video summarization in \DatasetName. We display the most common genres (13 classes) as in Moviescope dataset. Abbreviations: \textbf{Ac:} Action, \textbf{Ani:} Animation, \textbf{Bio:} Biography, \textbf{Com}: Comedy, \textbf{Cri}: Crime, \textbf{Drm}: Drama, \textbf{Fmy}: Family, \textbf{Fntsy}: Fantasy, \textbf{Hrrr}: Horror, \textbf{Myst}: Mystery, \textbf{Rom}: Romantic, \textbf{ScF}: SciFi, \textbf{Thrl}: Thriller. Other genre types including: History, Western, Adventure, Sports, Documentary, Music, Musical, and War.}  
\label{tab:sota}
\end{table*}




\subsection{Semantic analysis}

Semantic analysis is a crucial component in both unsupervised and supervised video summarization, as it greatly impacts the quality of the summary video. However, existing algorithms often overlook an important characteristic that is essential for all video summarization systems: the ability to handle unseen videos, also known as zero-shot capability. This means that the system should be able to perform well on videos that were not included in the training data, as this ultimately determines the system's quality. Taking inspiration from recent research demonstrating the generalization ability of large language models, we incorporate such a capability into our VSL architecture by employing semantic analysis at the natural language level.

Figure \ref{fig:pipeline}(b) illustrates the initial step in VSL's semantic analysis, which involves video captioning. In this phase, the video backbone of the input movie is first subsampled and then passed through a pre-trained BLIP model \cite{li2022blip} to generate captions for each sub-sampled frame. Following the video captioning stage, a multimodal summarization module is used. This module performs semantic summarization on the scene level for both video captioning and closed captioning, eliminating irrelevant information and enhancing the summarization performance. The final component of the semantic analysis is the scene scoring module, which uses a pre-trained T5 model \cite{2020t5} and cosine similarity. The T5 model generates embeddings for both the input genres and the outputs of the multimodal summarization module. If there are $L$ genres in the input genres, the score for scene $i$ of the input movie can be calculated using the following equation:

\begin{equation}\label{sceneScore}
    M_{i} = \sum_{l=0}^{L} (\frac{f^{vc}_{i} \cdot f_{g_{l}}}{\lVert f^{vc}_{i}\lVert  \lVert f_{g_{l}}\lVert} + \frac{f^{cc}_{i} \cdot f_{g_{l}}}{\lVert f^{cc}_{i}\lVert  \lVert f_{g_{l}}\lVert})
\end{equation}
where $f^{vc}$ is T5 feature for the scene's video captioning summarization, $f^{cc}$ is T5 feature for its closed captioning summarization.


\subsection{Video summarization frame selection}
The objective of video summarization is to select appropriate scenes from a list of scene scores $\{M_{1}, ..., M_{i}\}$ and include them in the final summary video. The length of the summary video can be adjusted based on user-specific settings (e.g., 10\%, 15\%, 25\% of the original movie). To handle different input videos, we offer two algorithms. For short movies (less than 5 minutes), the knapSack algorithm is employed to ensure diversity in the final summary video by balancing scene score and scene length during scene selection. 


\section{Experiment}

\subsection{Baselines}
\subsubsection{General video summarization.}

\noindent\textbf{CA-SUM} is a deep learning framework that employs a concentrated attention mechanism to evaluate the significance of frames in videos. 
The approach is unsupervised, eliminating the requirement for supervised data.
\noindent\textbf{RS-SUM} leverages self-supervised learning and a restorative score. The aim is to produce concise summaries from unannotated videos without the need for manual supervision. 
\noindent\textbf{AC-SUM-GAN} is an approach that combines an actor-critical model with a Generative Adversarial Network (GAN) to tackle the task of selecting important video fragments for creating a summary. This approach frames the selection process as a sequence generation task. 

\begin{table}[t]
\centering
\resizebox{0.48\textwidth}{!}{
\begin{tabular}{lcccc}
\toprule
\multirow{2}{*}{\textbf{Model}}& \textbf{Video Summarization}& \multicolumn{3}{c}{Genre number} \\
 \cmidrule(l){3-5}  
 & \textbf{Training Data}&1&2&3 \\
\midrule
MomentDETR \cite{momentdetr} &QV-highlight& 22.5 & 23.3 & 20.4 \\
QD-DETR \cite{qddetr} &QV-highlight& 23.1 & 23.2 & 23.4 \\
UniVTG \cite{lin2023univtg} &10 Datasets& 23.6 & 22.3& 20.1 \\
\midrule
\textbf{VSL} &- &\textbf{26.8} & \textbf{25.4} & \textbf{24.1}  \\
\bottomrule
\end{tabular}
}
\vspace{2mm}
\caption{Multi-genre video summarization on \DatasetName. Evaluated by F1 score. Our method outperform other baselines in all without any supervised training.}
\label{tab:multi_genre}
\vspace{-0.5cm}
\end{table}



\begin{table}[t]
\centering
\begin{tabular}{lccc}
\toprule
Model & Pre-trained data & Visual feature & F1 \\
\midrule
AC-SUM-GAN \cite{ac_sum} &  TVSum & GoogleNet  & 60.6 \\
CA-SUM \cite{CASum22}  &  TVSum & GoogleNet  & 61.4 \\
RS-SUM \cite{CASum22} &  TVSum & GoogleNet  & 61.4 \\
\midrule
\textbf{VSL} &  - & -  & \textbf{62.0}  \\
\bottomrule
\end{tabular}
\vspace{2mm}
\caption{General video summarization on TVSum. Our method doesn't require pre-training on the TVSum training data. Instead, we directly perform inferencing on the target data TVSum test split.}
\label{tab:general}
\end{table}

\begin{table}[t]
\centering
\begin{tabular}{lccc}
\toprule
Model & Pre-trained data & Visual feature & F1 \\
\midrule
MomentDETR \cite{momentdetr} &QV-highlight& CLIP  & 32.1 \\
QD-DETR \cite{qddetr} &QV-highlight& CLIP  & 31.5 \\
UniVTG \cite{lin2023univtg} &10 Datasets& CLIP  & 32.6 \\
\midrule
\textbf{VSL} &  - & -  & \textbf{34.8}  \\
\bottomrule
\end{tabular}
\vspace{2mm}
\caption{Summarization of user-generated videos on SumMe. We utilize the title as the query to generate summarization.}
\label{tab:summe}
\vspace{-0.3cm}
\end{table}


\subsubsection{Query-based video summarization} 

\noindent\textbf{MomentDETR} is the first end-to-end video summarization model that operates using a query-based approach. It adopts a transformer encoder-decoder architecture and treats video summarization as a direct set prediction task. 
Moreover, the model leverages the text query to assist in the selection of informative video segments, taking into account user preferences.
\noindent\textbf{QD-DETR} is built on top of MomentDETR and addresses the limited importance of queries in transformer architectures. In QD-DETR, the encoding module begins with cross-attention layers to incorporate the context of the text query into the video representation. 
\noindent\textbf{UniVTG} also adopts the end-to-end transformer architecture of MomentDETR. Research involves pre-training with a wide range of diverse labels obtained from different task-specific sources, which enhances its grounding capabilities.

\subsection{Experiment setup}
\noindent \textbf{VSL settings.}
In our VSL architecture test, we opt for a sample rate of 15 FPS for semantic analysis. The video captioning BLIP model is trained on the COCO dataset \cite{lin2014microsoft}. To generate closed captions, we employ the whisper-timestamped model \cite{lintoai2023whispertimestamped} to obtain more accurate timestamp results. During the test, we test out various genre types (21 classes in total), as indicated in Table \ref{tab:sota}. Additionally, we utilize the video summarization module with video-based scene detection results to enable comparison with our baseline algorithms. All experiments are conducted on Amazon Web Service, employing a single V100 GPU with 16GB of memory.

\noindent \textbf{Baseline settings.}
To generate video summarization, we utilize the UniVTG \cite{lin2023univtg} approach to pre-extract CLIP visual and text features from the query-based video summarization baselines. We employ the best checkpoints for each method, which have undergone finetuning or additional data pre-training, to test on the \DatasetName dataset. During the inference process, we input either a single genre or multiple genres as the text query to generate the video summarization.


\subsection{Evaluation metrics}
To comply with established protocols, we make sure that the summary generated using Keyshot is limited to less than 15\% of the original video duration, as stated in previous studies \cite{song2015tvsum}. To evaluate the accuracy of our predictions, we use the definitions of precision (P) and recall (R) as described in \cite{song2015tvsum}. Precision is calculated by dividing the duration of the overlapped prediction and ground-truth by the duration of the prediction, while recall is calculated by dividing the duration of the overlapped prediction and ground-truth by the duration of the ground-truth. The F-score is then calculated using the formula F = (2 x P x R)/(P + R) x 100\%. To ensure a reliable assessment, we employ a random selection approach. The F-score is reported for various experiment settings.

\subsection{User-preferred video summarization}


As indicated in Table \ref{tab:sota}, UniVTG achieves the highest performance among DETR-based methods by utilizing the largest training data compared to other methods. Our proposed method outperforms other baselines trained on video summarization datasets such as QV-highlight and TVSum. By combining the captioning model for video understanding with the summarization model, our approach effectively captures the semantic information of videos across multiple genre-related scenes. Importantly, this behavior is learned without the need for additional human supervision or labeled data. In contrast, the current DETR-based method is trained on natural language descriptions as queries, requiring additional fine-tuning to adapt well to video summarization with specific genres.



\begin{figure}
\subfigure[\textbf{Scaling \# of videos.} Each video contains 1-3 movie genres. VSL achieved the lowest runtime among all methods.]
{\includegraphics[width=0.48\linewidth]{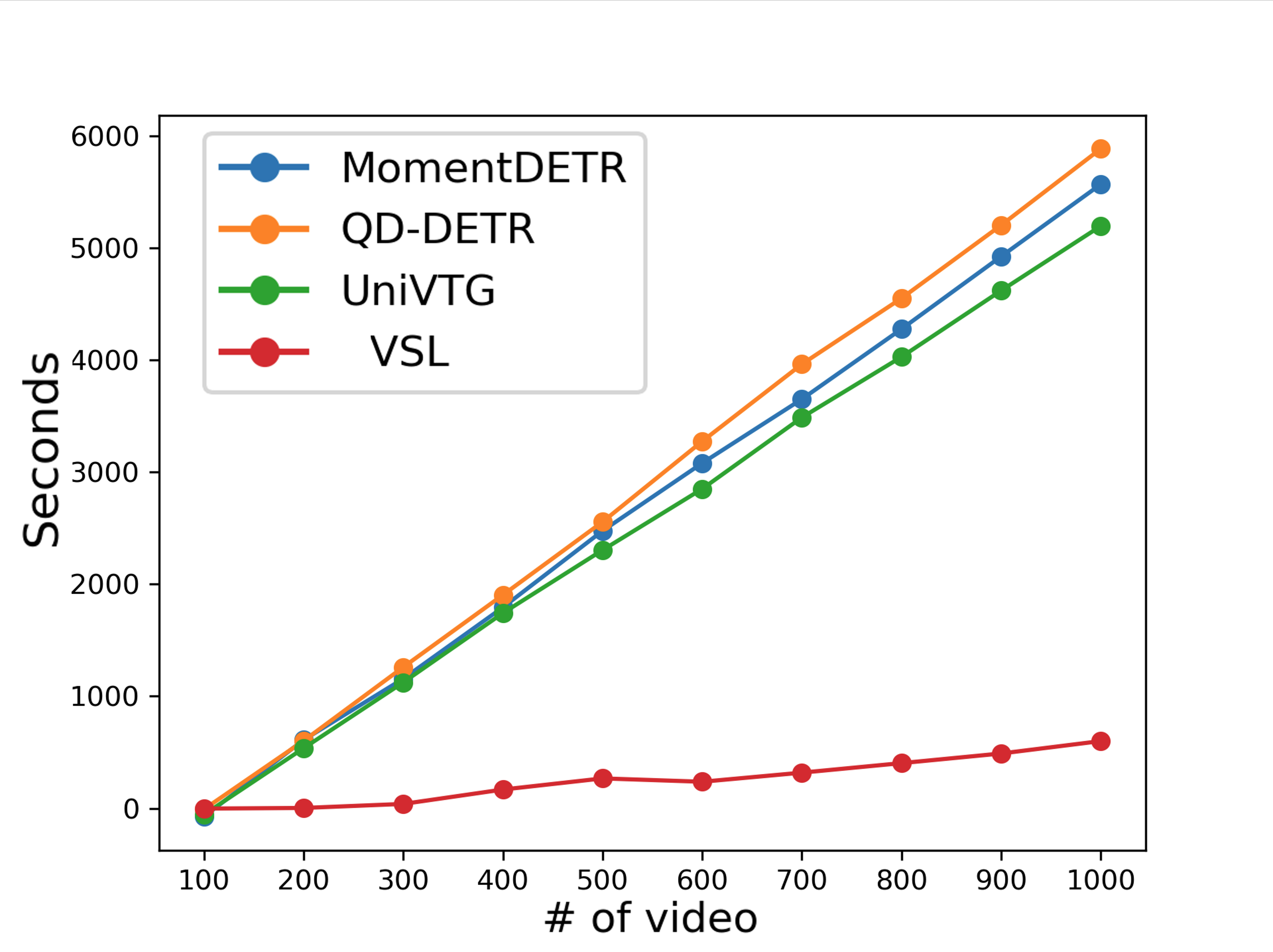}}
\hfill
\subfigure[\textbf{Scaling \# of user preferences.} We fix the video number to 10. VSL achieved constant time while the number of user preferences scaled up.]{\includegraphics[width=0.48\linewidth]{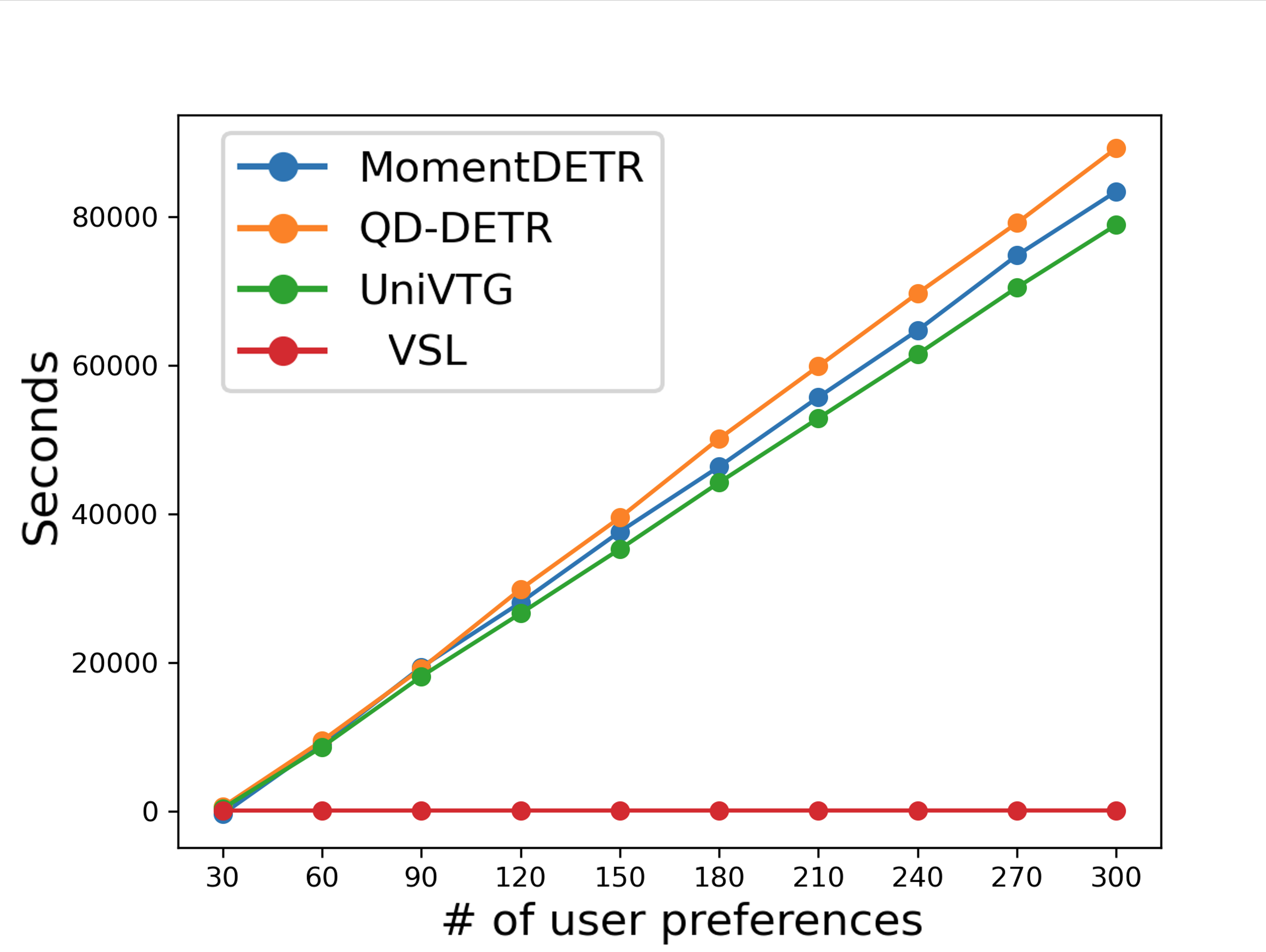}}
\vspace{-0.3cm}
\caption{Runtime analysis of \# of videos and preferences.}
\vspace{-0.1cm}
\label{fig:runtime}
\end{figure}

\subsection{Multi-genre video summarization.}
\label{sec:more_ex_multi}
We evaluate the performance with different \# of genres as in Table \ref{tab:multi_genre}. In this setting, a user is interested in multiple genres, such as Sci-Fi and Romance. To obtain annotations for such cases, we follow the label generation process described in Figure \ref{fig:dataset}(e). We aggregate the confidence scores in the $N$ genres and apply the same filtering criteria. In the experiment, we observe that MomentDETR and UniVTG experience a significant drop in performance as the number of genres increases ($N$=3). In contrast, QD-DETR demonstrates consistent performance across different numbers of genres. We attribute this difference to the unique design of QD-DETR, which incorporates additional interaction between video and text. 
Our proposed method consistently outperforms the current baselines, without requiring video summarization label training. This highlights the effectiveness and generalizability of our approach.

\begin{table}[t]
        \centering
		 \begin{tabularx}{.45\textwidth}{lp{2em}c@{\hskip 0.3in}c}
			\toprule
				\multirow{1}{*}{\textbf{Text}} & & Freqeunt Caption   & Summarization   \\
				\hline
				F1                          &  &   31.4 & 34.8 \\
			\bottomrule
		\end{tabularx}

        \bigskip
        \begin{tabularx}{.45\textwidth}{lp{1em}c@{\hskip 0.3in}c@{\hskip 0.3in}c@{\hskip 0.3in}c}
				\multirow{1}{*}{\textbf{LM}} & & Glove   & T5-base  & T5-large& T5-XL \\
				\hline
				F1                          &  &   25.5 & 22.7 &22.9 &26.8 \\
			\bottomrule
		\end{tabularx}
		\vspace{0.2cm}
		 \caption{
        \textbf{Ablations study.} 
        \textbf{(a)} The summarization captured useful information from the diverse scenes leads to better performance.
        \textbf{(b)} We test out various language models (LM) for the semantic matching component.
    }
    \label{tab:ablation}
    \vspace{-0.9cm}
\end{table}

\subsection{Evaluation on public benchmarks} 
\noindent \textbf{TV shows.}
We evaluate our approach by comparing it with existing unsupervised methods on the TVSum dataset, as presented in Table \ref{tab:general}. To conduct the experiment, we utilize the video genre information from the metadata of the TVSum dataset and set the summary video length ratio to 0.15. 
The VSL method outperforms the other unsupervised baselines, indicating its effectiveness and versatility on a standard video summarization dataset.

\noindent \textbf{User generated videos.}
We test on a user-generated video dataset SumMe\cite{summe}, where live videos are generated by users. 
As in Table \ref{tab:summe}, we conducted a comparison with other query-based video summarization methods, using the video title as the query. The results showed that our VSL method performed satisfactorily in comparison to other baselines, thereby proving its adaptability to a variety of video types, not limited to movies or TV shows.


\subsection{Ablation study}
We've conducted an ablation study on two elements of our model, as shown in Table \ref{tab:ablation}. 

\noindent \textbf{Multimodal summarization.} We abate the component depicted in Figure \ref{fig:pipeline} (b) (Multimodal summarization) by substituting it with the caption that appears most frequently in the videos, rather than employing the summarization model. 
Results in the SumMe dataset show the efficacy of the summarization model, which encompasses a more thorough understanding of the full length of the video.

\noindent \textbf{Language Model (LM) for semantic matching.}
We tested various language models in Figure \ref{fig:pipeline} (b) (Scene scoring)  on our \DatasetName~ dataset.
Our findings indicate that basic word embedding (Glove) semantic matching results in satisfactory performance. Moreover, training language models with a greater number of parameters enhances semantic comprehension, which is beneficial for everything from summarization to genre matching.




\subsection{Runtime analysis} 
\noindent \textbf{Scaling \# of videos.} In order to test the practical application of video summarization, we assume that the user's preferences have been obtained in advance. Our goal is to generate video summarization in real-time. The runtime of each method is calculated and presented in Figure \ref{fig:runtime}(a). It was observed that QD-DETR requires the longest inference time due to its complex architecture, which includes a specially designed cross-attention mechanism between queries and videos. Unlike DETR-based methods, VSL is able to maintain a similar runtime as the number of input videos increases. This is because our method converts the video into text beforehand, allowing for faster processing of video summarization selection. In other words, we avoid the overhead of video processing to generate video summarization more quickly. 

\noindent \textbf{Scaling \# of user preferences.}
In practical scenarios, users may have various combinations of genre preferences. In our specific case, there are 21 different genres available and each user can have a maximum of 3 preferred genres. Among these combinations, we selected 300 combinations of user preferences across 10 videos, as depicted in Figure \ref{fig:runtime}(b). Although the DETR-based method requires feeding individual queries to the model, our model can precompute genre features, ensuring a constant time complexity as the number of user-preference combinations increases.


\section{Conclusion}

In this study, we have introduced a new dataset and a pipeline for multimodal video summarization. Our approach utilizes pre-trained visual language models and semantic analysis at the scene level, which eliminates the need for extensive training datasets. The experimental results demonstrate the effectiveness of our approach, surpassing existing unsupervised video summarization models, and demonstrating its adaptability across different datasets. This represents a significant advance in making video summarization more accessible, efficient, and customizable to individual user preferences. 
In the end, our method is not confined to genre; genre is merely one mechanism to encapsulate user preferences, but any other technique that captures user preferences can be utilized.
As a future direction, 
we plan to experiment with new methods of expressing user preferences, such as generating user profiles based on LLM.


{
    \small
    \bibliographystyle{ACM-Reference-Format}
    \balance 
    \bibliography{sample-base}


\begin{thebibliography}{29}


\ifx \showCODEN    \undefined \def \showCODEN     #1{\unskip}     \fi
\ifx \showDOI      \undefined \def \showDOI       #1{#1}\fi
\ifx \showISBNx    \undefined \def \showISBNx     #1{\unskip}     \fi
\ifx \showISBNxiii \undefined \def \showISBNxiii  #1{\unskip}     \fi
\ifx \showISSN     \undefined \def \showISSN      #1{\unskip}     \fi
\ifx \showLCCN     \undefined \def \showLCCN      #1{\unskip}     \fi
\ifx \shownote     \undefined \def \shownote      #1{#1}          \fi
\ifx \showarticletitle \undefined \def \showarticletitle #1{#1}   \fi
\ifx \showURL      \undefined \def \showURL       {\relax}        \fi
\providecommand\bibfield[2]{#2}
\providecommand\bibinfo[2]{#2}
\providecommand\natexlab[1]{#1}
\providecommand\showeprint[2][]{arXiv:#2}

\bibitem[Apostolidis et~al\mbox{.}(2020)]%
        {ac_sum}
\bibfield{author}{\bibinfo{person}{Evlampios Apostolidis}, \bibinfo{person}{Eleni Adamantidou}, \bibinfo{person}{Alexandros~I Metsai}, \bibinfo{person}{Vasileios Mezaris}, {and} \bibinfo{person}{Ioannis Patras}.} \bibinfo{year}{2020}\natexlab{}.
\newblock \showarticletitle{AC-SUM-GAN: Connecting actor-critic and generative adversarial networks for unsupervised video summarization}.
\newblock \bibinfo{journal}{\emph{IEEE Transactions on Circuits and Systems for Video Technology}} \bibinfo{volume}{31}, \bibinfo{number}{8} (\bibinfo{year}{2020}), \bibinfo{pages}{3278--3292}.
\newblock


\bibitem[Apostolidis et~al\mbox{.}(2022)]%
        {CASum22}
\bibfield{author}{\bibinfo{person}{Evlampios Apostolidis}, \bibinfo{person}{Georgios Balaouras}, \bibinfo{person}{Vasileios Mezaris}, {and} \bibinfo{person}{Ioannis Patras}.} \bibinfo{year}{2022}\natexlab{}.
\newblock \showarticletitle{Summarizing Videos Using Concentrated Attention and Considering the Uniqueness and Diversity of the Video Frames} \emph{(\bibinfo{series}{ICMR '22})}. \bibinfo{publisher}{Association for Computing Machinery}, \bibinfo{address}{New York, NY, USA}, \bibinfo{pages}{407--415}.
\newblock
\showISBNx{9781450392389}
\urldef\tempurl%
\url{https://doi.org/10.1145/3512527.3531404}
\showDOI{\tempurl}


\bibitem[Artstein(2017)]%
        {artstein2017inter}
\bibfield{author}{\bibinfo{person}{Ron Artstein}.} \bibinfo{year}{2017}\natexlab{}.
\newblock \showarticletitle{Inter-annotator agreement}.
\newblock \bibinfo{journal}{\emph{Handbook of linguistic annotation}} (\bibinfo{year}{2017}), \bibinfo{pages}{297--313}.
\newblock


\bibitem[Bain et~al\mbox{.}(2020)]%
        {condensed_movie}
\bibfield{author}{\bibinfo{person}{Max Bain}, \bibinfo{person}{Arsha Nagrani}, \bibinfo{person}{Andrew Brown}, {and} \bibinfo{person}{Andrew Zisserman}.} \bibinfo{year}{2020}\natexlab{}.
\newblock \showarticletitle{Condensed movies: Story based retrieval with contextual embeddings}. In \bibinfo{booktitle}{\emph{Proceedings of the Asian Conference on Computer Vision}}.
\newblock


\bibitem[Bose et~al\mbox{.}(2023)]%
        {bose2023movieclip}
\bibfield{author}{\bibinfo{person}{Digbalay Bose}, \bibinfo{person}{Rajat Hebbar}, \bibinfo{person}{Krishna Somandepalli}, \bibinfo{person}{Haoyang Zhang}, \bibinfo{person}{Yin Cui}, \bibinfo{person}{Kree Cole-McLaughlin}, \bibinfo{person}{Huisheng Wang}, {and} \bibinfo{person}{Shrikanth Narayanan}.} \bibinfo{year}{2023}\natexlab{}.
\newblock \showarticletitle{Movieclip: Visual scene recognition in movies}. In \bibinfo{booktitle}{\emph{Proceedings of the IEEE/CVF Winter Conference on Applications of Computer Vision}}. \bibinfo{pages}{2083--2092}.
\newblock


\bibitem[Cascante-Bonilla et~al\mbox{.}(2019)]%
        {2019Moviescope}
\bibfield{author}{\bibinfo{person}{Paola Cascante-Bonilla}, \bibinfo{person}{Kalpathy Sitaraman}, \bibinfo{person}{Mengjia Luo}, {and} \bibinfo{person}{Vicente Ordonez}.} \bibinfo{year}{2019}\natexlab{}.
\newblock \showarticletitle{Moviescope: Large-scale Analysis of Movies using Multiple Modalities}.
\newblock \bibinfo{journal}{\emph{ArXiv}}  \bibinfo{volume}{abs/1908.03180} (\bibinfo{year}{2019}).
\newblock


\bibitem[Gygli et~al\mbox{.}(2014)]%
        {summe}
\bibfield{author}{\bibinfo{person}{Michael Gygli}, \bibinfo{person}{Helmut Grabner}, \bibinfo{person}{Hayko Riemenschneider}, {and} \bibinfo{person}{Luc~Van Gool}.} \bibinfo{year}{2014}\natexlab{}.
\newblock \showarticletitle{Creating summaries from user videos}. In \bibinfo{booktitle}{\emph{ECCV}}. Springer, \bibinfo{pages}{505--520}.
\newblock


\bibitem[Huang and Worring(2020)]%
        {huang2020query}
\bibfield{author}{\bibinfo{person}{Jia-Hong Huang} {and} \bibinfo{person}{Marcel Worring}.} \bibinfo{year}{2020}\natexlab{}.
\newblock \showarticletitle{Query-controllable video summarization}. In \bibinfo{booktitle}{\emph{Proceedings of the 2020 International Conference on Multimedia Retrieval}}. \bibinfo{pages}{242--250}.
\newblock


\bibitem[Jiang and Mu(2022)]%
        {jiang2022joint}
\bibfield{author}{\bibinfo{person}{Hao Jiang} {and} \bibinfo{person}{Yadong Mu}.} \bibinfo{year}{2022}\natexlab{}.
\newblock \showarticletitle{Joint Video Summarization and Moment Localization by Cross-Task Sample Transfer}. In \bibinfo{booktitle}{\emph{CVPR}}. \bibinfo{pages}{16388--16398}.
\newblock


\bibitem[Lei et~al\mbox{.}(2021)]%
        {momentdetr}
\bibfield{author}{\bibinfo{person}{Jie Lei}, \bibinfo{person}{Tamara~L Berg}, {and} \bibinfo{person}{Mohit Bansal}.} \bibinfo{year}{2021}\natexlab{}.
\newblock \showarticletitle{Detecting moments and highlights in videos via natural language queries}.
\newblock \bibinfo{journal}{\emph{Advances in Neural Information Processing Systems}}  \bibinfo{volume}{34} (\bibinfo{year}{2021}), \bibinfo{pages}{11846--11858}.
\newblock


\bibitem[Li et~al\mbox{.}(2022)]%
        {li2022blip}
\bibfield{author}{\bibinfo{person}{Junnan Li}, \bibinfo{person}{Dongxu Li}, \bibinfo{person}{Caiming Xiong}, {and} \bibinfo{person}{Steven Hoi}.} \bibinfo{year}{2022}\natexlab{}.
\newblock \showarticletitle{Blip: Bootstrapping language-image pre-training for unified vision-language understanding and generation}. In \bibinfo{booktitle}{\emph{International Conference on Machine Learning}}. PMLR, \bibinfo{pages}{12888--12900}.
\newblock


\bibitem[Lin et~al\mbox{.}(2023)]%
        {lin2023univtg}
\bibfield{author}{\bibinfo{person}{Kevin~Qinghong Lin}, \bibinfo{person}{Pengchuan Zhang}, \bibinfo{person}{Joya Chen}, \bibinfo{person}{Shraman Pramanick}, \bibinfo{person}{Difei Gao}, \bibinfo{person}{Alex~Jinpeng Wang}, \bibinfo{person}{Rui Yan}, {and} \bibinfo{person}{Mike~Zheng Shou}.} \bibinfo{year}{2023}\natexlab{}.
\newblock \showarticletitle{UniVTG: Towards Unified Video-Language Temporal Grounding}. In \bibinfo{booktitle}{\emph{Proceedings of the IEEE/CVF International Conference on Computer Vision}}. \bibinfo{pages}{2794--2804}.
\newblock


\bibitem[Lin et~al\mbox{.}(2014)]%
        {lin2014microsoft}
\bibfield{author}{\bibinfo{person}{Tsung-Yi Lin}, \bibinfo{person}{Michael Maire}, \bibinfo{person}{Serge Belongie}, \bibinfo{person}{James Hays}, \bibinfo{person}{Pietro Perona}, \bibinfo{person}{Deva Ramanan}, \bibinfo{person}{Piotr Doll{\'a}r}, {and} \bibinfo{person}{C~Lawrence Zitnick}.} \bibinfo{year}{2014}\natexlab{}.
\newblock \showarticletitle{Microsoft coco: Common objects in context}. In \bibinfo{booktitle}{\emph{Computer Vision--ECCV 2014: 13th European Conference, Zurich, Switzerland, September 6-12, 2014, Proceedings, Part V 13}}. Springer, \bibinfo{pages}{740--755}.
\newblock


\bibitem[Louradour(2023)]%
        {lintoai2023whispertimestamped}
\bibfield{author}{\bibinfo{person}{J{\'e}r{\^o}me Louradour}.} \bibinfo{year}{2023}\natexlab{}.
\newblock \bibinfo{title}{whisper-timestamped}.
\newblock \bibinfo{howpublished}{\url{https://github.com/linto-ai/whisper-timestamped}}.
\newblock


\bibitem[Luo et~al\mbox{.}(2022)]%
        {luo2022clip4clip}
\bibfield{author}{\bibinfo{person}{Huaishao Luo}, \bibinfo{person}{Lei Ji}, \bibinfo{person}{Ming Zhong}, \bibinfo{person}{Yang Chen}, \bibinfo{person}{Wen Lei}, \bibinfo{person}{Nan Duan}, {and} \bibinfo{person}{Tianrui Li}.} \bibinfo{year}{2022}\natexlab{}.
\newblock \showarticletitle{Clip4clip: An empirical study of clip for end to end video clip retrieval and captioning}.
\newblock \bibinfo{journal}{\emph{Neurocomputing}}  \bibinfo{volume}{508} (\bibinfo{year}{2022}), \bibinfo{pages}{293--304}.
\newblock


\bibitem[Mahasseni et~al\mbox{.}(2017)]%
        {mahasseni2017unsupervised}
\bibfield{author}{\bibinfo{person}{Behrooz Mahasseni}, \bibinfo{person}{Michael Lam}, {and} \bibinfo{person}{Sinisa Todorovic}.} \bibinfo{year}{2017}\natexlab{}.
\newblock \showarticletitle{Unsupervised video summarization with adversarial lstm networks}. In \bibinfo{booktitle}{\emph{CVPR}}. \bibinfo{pages}{202--211}.
\newblock


\bibitem[Moon et~al\mbox{.}(2023)]%
        {qddetr}
\bibfield{author}{\bibinfo{person}{WonJun Moon}, \bibinfo{person}{Sangeek Hyun}, \bibinfo{person}{SangUk Park}, \bibinfo{person}{Dongchan Park}, {and} \bibinfo{person}{Jae-Pil Heo}.} \bibinfo{year}{2023}\natexlab{}.
\newblock \showarticletitle{Query-dependent video representation for moment retrieval and highlight detection}. In \bibinfo{booktitle}{\emph{Proceedings of the IEEE/CVF Conference on Computer Vision and Pattern Recognition}}. \bibinfo{pages}{23023--23033}.
\newblock


\bibitem[Nalla et~al\mbox{.}(2020)]%
        {nalla2020watch}
\bibfield{author}{\bibinfo{person}{Saiteja Nalla}, \bibinfo{person}{Mohit Agrawal}, \bibinfo{person}{Vishal Kaushal}, \bibinfo{person}{Ganesh Ramakrishnan}, {and} \bibinfo{person}{Rishabh Iyer}.} \bibinfo{year}{2020}\natexlab{}.
\newblock \showarticletitle{Watch Hours in Minutes: Summarizing Videos with User Intent}. In \bibinfo{booktitle}{\emph{ECCV}}.
\newblock


\bibitem[Narasimhan et~al\mbox{.}(2021)]%
        {narasimhan2021clip}
\bibfield{author}{\bibinfo{person}{Medhini Narasimhan}, \bibinfo{person}{Anna Rohrbach}, {and} \bibinfo{person}{Trevor Darrell}.} \bibinfo{year}{2021}\natexlab{}.
\newblock \showarticletitle{CLIP-It! language-guided video summarization}.
\newblock \bibinfo{journal}{\emph{NeurIPS}}, \bibinfo{pages}{13988--14000}.
\newblock


\bibitem[Pazzani and Billsus(2007)]%
        {pazzani2007content}
\bibfield{author}{\bibinfo{person}{Michael~J Pazzani} {and} \bibinfo{person}{Daniel Billsus}.} \bibinfo{year}{2007}\natexlab{}.
\newblock \showarticletitle{Content-based recommendation systems}.
\newblock In \bibinfo{booktitle}{\emph{The adaptive web: methods and strategies of web personalization}}. \bibinfo{publisher}{Springer}, \bibinfo{pages}{325--341}.
\newblock


\bibitem[Radford et~al\mbox{.}(2021)]%
        {CLIP}
\bibfield{author}{\bibinfo{person}{Alec Radford}, \bibinfo{person}{Jong~Wook Kim}, \bibinfo{person}{Chris Hallacy}, \bibinfo{person}{Aditya Ramesh}, \bibinfo{person}{Gabriel Goh}, \bibinfo{person}{Sandhini Agarwal}, \bibinfo{person}{Girish Sastry}, \bibinfo{person}{Amanda Askell}, \bibinfo{person}{Pamela Mishkin}, \bibinfo{person}{Jack Clark}, \bibinfo{person}{Gretchen Krueger}, {and} \bibinfo{person}{Ilya Sutskever}.} \bibinfo{year}{2021}\natexlab{}.
\newblock \showarticletitle{Learning Transferable Visual Models From Natural Language Supervision}.
\newblock \bibinfo{journal}{\emph{CoRR}}  \bibinfo{volume}{abs/2103.00020} (\bibinfo{year}{2021}).
\newblock
\showeprint[arXiv]{2103.00020}
\urldef\tempurl%
\url{https://arxiv.org/abs/2103.00020}
\showURL{%
\tempurl}


\bibitem[Raffel et~al\mbox{.}(2020)]%
        {2020t5}
\bibfield{author}{\bibinfo{person}{Colin Raffel}, \bibinfo{person}{Noam Shazeer}, \bibinfo{person}{Adam Roberts}, \bibinfo{person}{Katherine Lee}, \bibinfo{person}{Sharan Narang}, \bibinfo{person}{Michael Matena}, \bibinfo{person}{Yanqi Zhou}, \bibinfo{person}{Wei Li}, {and} \bibinfo{person}{Peter~J. Liu}.} \bibinfo{year}{2020}\natexlab{}.
\newblock \showarticletitle{Exploring the Limits of Transfer Learning with a Unified Text-to-Text Transformer}.
\newblock \bibinfo{journal}{\emph{Journal of Machine Learning Research}} \bibinfo{volume}{21}, \bibinfo{number}{140} (\bibinfo{year}{2020}), \bibinfo{pages}{1--67}.
\newblock
\urldef\tempurl%
\url{http://jmlr.org/papers/v21/20-074.html}
\showURL{%
\tempurl}


\bibitem[Sharghi et~al\mbox{.}(2017)]%
        {sharghi2017query}
\bibfield{author}{\bibinfo{person}{Aidean Sharghi}, \bibinfo{person}{Jacob~S Laurel}, {and} \bibinfo{person}{Boqing Gong}.} \bibinfo{year}{2017}\natexlab{}.
\newblock \showarticletitle{Query-focused video summarization: Dataset, evaluation, and a memory network based approach}. In \bibinfo{booktitle}{\emph{CVPR}}.
\newblock


\bibitem[Song et~al\mbox{.}(2015)]%
        {song2015tvsum}
\bibfield{author}{\bibinfo{person}{Yale Song}, \bibinfo{person}{Jordi Vallmitjana}, \bibinfo{person}{Amanda Stent}, {and} \bibinfo{person}{Alejandro Jaimes}.} \bibinfo{year}{2015}\natexlab{}.
\newblock \showarticletitle{Tvsum: Summarizing web videos using titles}. In \bibinfo{booktitle}{\emph{Proceedings of the IEEE conference on computer vision and pattern recognition}}. \bibinfo{pages}{5179--5187}.
\newblock


\bibitem[Vasudevan et~al\mbox{.}(2017)]%
        {vasudevan2017query}
\bibfield{author}{\bibinfo{person}{Arun~Balajee Vasudevan}, \bibinfo{person}{Michael Gygli}, \bibinfo{person}{Anna Volokitin}, {and} \bibinfo{person}{Luc Van~Gool}.} \bibinfo{year}{2017}\natexlab{}.
\newblock \showarticletitle{Query-adaptive video summarization via quality-aware relevance estimation}. In \bibinfo{booktitle}{\emph{ACM MM}}.
\newblock


\bibitem[Wu et~al\mbox{.}(2022)]%
        {wu2022intentvizor}
\bibfield{author}{\bibinfo{person}{Guande Wu}, \bibinfo{person}{Jianzhe Lin}, {and} \bibinfo{person}{Claudio~T Silva}.} \bibinfo{year}{2022}\natexlab{}.
\newblock \showarticletitle{IntentVizor: Towards Generic Query Guided Interactive Video Summarization}. In \bibinfo{booktitle}{\emph{CVPR}}. \bibinfo{pages}{10503--10512}.
\newblock


\bibitem[Xiao et~al\mbox{.}(2020a)]%
        {xiao2020query}
\bibfield{author}{\bibinfo{person}{Shuwen Xiao}, \bibinfo{person}{Zhou Zhao}, \bibinfo{person}{Zijian Zhang}, \bibinfo{person}{Ziyu Guan}, {and} \bibinfo{person}{Deng Cai}.} \bibinfo{year}{2020}\natexlab{a}.
\newblock \showarticletitle{Query-biased self-attentive network for query-focused video summarization}.
\newblock \bibinfo{journal}{\emph{{IEEE} Trans. Image Process.}}  \bibinfo{volume}{29} (\bibinfo{year}{2020}), \bibinfo{pages}{5889--5899}.
\newblock


\bibitem[Xiao et~al\mbox{.}(2020b)]%
        {xiao2020convolutional}
\bibfield{author}{\bibinfo{person}{Shuwen Xiao}, \bibinfo{person}{Zhou Zhao}, \bibinfo{person}{Zijian Zhang}, \bibinfo{person}{Xiaohui Yan}, {and} \bibinfo{person}{Min Yang}.} \bibinfo{year}{2020}\natexlab{b}.
\newblock \showarticletitle{Convolutional hierarchical attention network for query-focused video summarization}. In \bibinfo{booktitle}{\emph{AAAI}}, Vol.~\bibinfo{volume}{34}. \bibinfo{pages}{12426--12433}.
\newblock


\bibitem[Zeng et~al\mbox{.}(2022)]%
        {zeng2022socratic}
\bibfield{author}{\bibinfo{person}{Andy Zeng}, \bibinfo{person}{Maria Attarian}, \bibinfo{person}{Brian Ichter}, \bibinfo{person}{Krzysztof Choromanski}, \bibinfo{person}{Adrian Wong}, \bibinfo{person}{Stefan Welker}, \bibinfo{person}{Federico Tombari}, \bibinfo{person}{Aveek Purohit}, \bibinfo{person}{Michael Ryoo}, \bibinfo{person}{Vikas Sindhwani}, {et~al\mbox{.}}} \bibinfo{year}{2022}\natexlab{}.
\newblock \showarticletitle{Socratic models: Composing zero-shot multimodal reasoning with language}.
\newblock \bibinfo{journal}{\emph{arXiv preprint arXiv:2204.00598}} (\bibinfo{year}{2022}).
\newblock


\end{thebibliography}
}
\end{document}